\documentclass[10pt,twocolumn,letterpaper]{article}

\usepackage[pagenumbers]{wacv} % To force page numbers, e.g. for an arXiv version

\usepackage{graphicx}
\usepackage{amsmath}
\usepackage{amssymb}
\usepackage{booktabs}

\usepackage[pagebackref,breaklinks,colorlinks]{hyperref}

\usepackage{tabularx}
\usepackage{array, float, multirow}
\usepackage{booktabs,caption}
\usepackage[capitalize]{cleveref}
\crefname{section}{Sec.}{Secs.}
\Crefname{section}{Section}{Sections}
\Crefname{table}{Table}{Tables}
\crefname{table}{Tab.}{Tabs.}

\def\wacvPaperID{379} % *** Enter the WACV Paper ID here
\def\confName{WACV}
\def\confYear{2024}

\begin{document}

\title{Squeeze aggregated excitation network}

\author{Mahendran N\\
{\tt\small mahendranNNM@gmail.com}
}
\maketitle

%%%%%%%%% ABSTRACT
\begin{abstract}
   Convolutional neural networks have spatial representations which read patterns in the vision tasks. Squeeze and excitation links the channel wise representations by explicitly modeling on channel level. Multi layer perceptrons learn global representations and in most of the models it is used often at the end after all convolutional layers to gather all the information learned before classification. We propose a method of inducing the global representations within channels to have better performance of the model. We propose SaEnet, Squeeze aggregated excitation network, for learning global channelwise representation in between layers. The proposed module takes advantage of passing important information after squeeze by having aggregated excitation before regaining its shape. We also introduce a new idea of having a multibranch linear(dense) layer in the network. This learns global representations from the condensed information which enhances the representational power of the network. The proposed module have undergone extensive experiments by using Imagenet and CIFAR100 datasets and compared with closely related architectures. The analyzes results that proposed models outputs are comparable and in some cases better than existing state of the art architectures.
\end{abstract}

%%%%%%%%% BODY TEXT
\section{Introduction}
\label{sec:intro}

Convolutional neural networks (CNNs) have always been emerging in network engineering irrespective of the current trends. Multiple architectures have been proposed by identifying problems and providing solutions one-by-one \cite{inception,resnet1000}. Various architectures emerged after the success of AlexNet \cite{alexnet}, including VGG \cite{vgg2014}, Residual network \cite{resnet}. Convolution based architectures have been exploring to increase the representational power of the network \cite{senet}. Researchers have been trying to enhance the representations of deep learning architectures in various ways as CNN is effective for vision based tasks. There are multiple representations with which layers of neural networks learn for vision tasks. 

Convolutional networks learn with spatial correlations within local receptive field. Spatial correlations are important for learning new features for performing models. CNN learns the spatial representations which is key for identifying patterns in images. Fully connected (FC) layers learns global representations as they have connections with all the nodes in that layer. Fully connected layers are often used in architectures at the nearing end as it helps in classification of results using Softmax layer. Squeeze and Excitation network (SENet), one of the latest developments in this field, have explicitly modeling interdependencies of channels \cite{senet}. This enhances the representational power of the network by feature recalibration. SENet \cite{senet} made changes internally within channels to have a impact on performance of the model. This module selectively sends important information to the next layer from globally learned information. Squeeze and excitation network proposed to have channelwise representations in the model.

%[Inception points]
Christian et al. \cite{inception} proposed Inception module which introduces a new architectural design of having an optimal sparse CNN within the block. The multibranch convolutions consists of different filter sizes which are concatenated at the end of the block. This new strategy is ignites the new architectural topology and thereby proposing a method of achieving better performance by having less theoretical complexity. The inception module usually contains convolutional layers to make it learn spatial representation. Similar approaches have been proposed like resnext which have multiple branches of same topology within module. These aggregated modules is also known to be measured with new dimension name called as cardinality. Aggregated networks make network learn more spatial correlations without increasing depth and utilizing the computation effectively. Further researchers have adopted this inception based approach and built network architectures \cite{resnext,xception,biglittlenet,wrns,mobilenet}.
%The recent researches have made the addition of learning mechanisms explicitly for enhancing accuracy [Inception, BN(in inception)]. 

\begin{figure*}[hbt!]
    \centering
    \includegraphics[width=\textwidth,height=10cm]{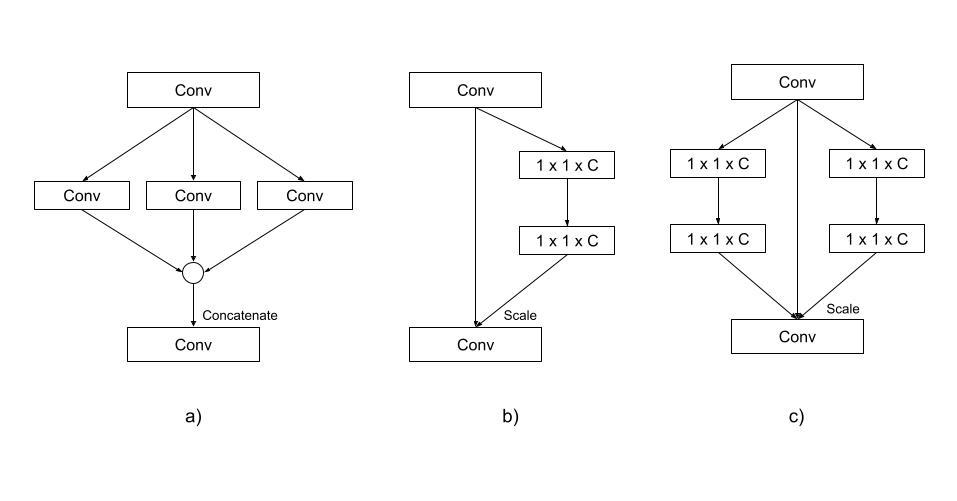}
    \caption{The comparison between the resnext, seresnet and proposed saeresnet modules are shown. a) The aggregated module of multibranch convolutional layers having less theoretical complexity. b) The squeeze and excitation module consists of reduction of input size to 1 along their channels and then regaining its original shape. Input to squeeze operation is sent to regain the original shape unlike aggregated network.  c)The proposed SaEnet module containing a mix of a) and b). The squeeze performs reduction of input and the reduced input is passed to aggregated layers for learning more representations than the SEnet. Results are concatenated and follows the SEnet for regaining their shape.}
    \label{fig:mainimages}
% \centering
\end{figure*}
%SEnet

%end senet

Fully connected layers learns global representations from all the previous layers and use it for classification. These learned have been used towards the end in usual classification architectures. Squeeze and excitation networks \cite{senet} proposes channelwise representation learning with the help of FC layers. SEnet makes channelwise representatiosn stronger by squeeze and excitation operations. Channel wise features are recalibrated with the squeezed input. The excitation part of proposed module consists of fully connected layer which learns global representation. This module make sure that only the most important features are passed to the next layer. This specific module design when repeatedly called in networks like ResNet inside the residual module can have a greater impact as it acts as a filter to the network.

\begin{figure}
    \centering
    \includegraphics[width=0.9\linewidth,height=10cm]{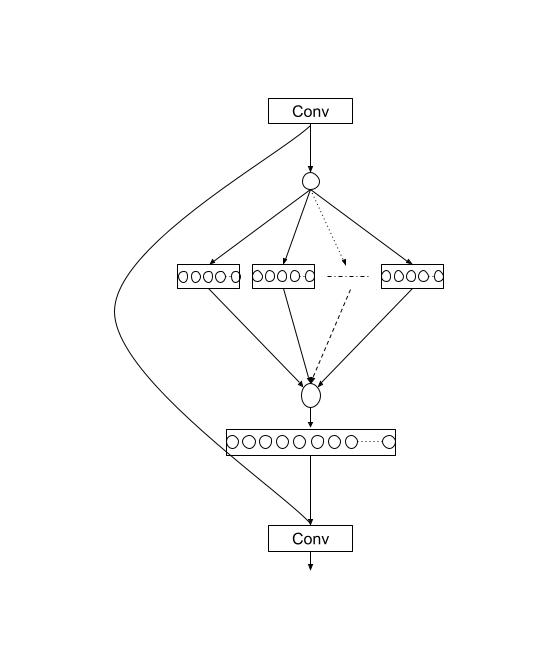}
    \caption{The figure depicts the inside working of proposed SaE module. The inputs from convolution is being squeezed then passed to aggregated FC layers which follows the excitation process. The splitted input is passed at the end to regain the original shape for future layers.}
    \label{fig:complete}
\end{figure}

The comparison between the existing aggregated residual module, the Squeeze and excitation module and the proposed squeeze aggregated excitation module is shown in Figure \ref{fig:mainimages}. From the figure, the Squeeze excitation module passes important features in both SE and SaE module. In the proposed SaE module, we utilize this stage of module in a better way by increasing cardinality between layers. We have increased cardinality of the first FC layer after the squeeze module. From the inspiration of inception module we use aggregated FC layer of same size similar to resnext. Since this has increased impact than stacked layers as discussed \cite{resnext}. This not only makes important features to be learned by global representations in module but also have better performance with increased cardinality. The proposed module is shown in Figure \ref{fig:complete}. The proposed module have better theoretical complexity than existing SEmodule. We use a reduction size of 32 and cardinality of 4. We keep the cardinality values small as the important features are being learned and not to increase the complexity. The results from the aggregated FC layers are concatenated in the excitation phase and regained its output shape to pass it to upcoming layers. The entire operation of Squeeze aggregated excitaion module containing the aggregated FC layers is shown in Figure \ref{fig:complete}

In this paper, we propose a combination of architectural designs by linking spatial, channel wise  and global representations. We brought a mix of SEnet with aggregated resnet to propose this module, SaEnet, Squeeze aggregated excitation module.

The main contributions of this paper is summarized as follows:
\begin{itemize}
    \item We recognize the impact of the aggregated modules and representational power from Inception architectures and squeeze and excitation module respectively. Aggregated module reduce the theoretical complexity of the network.
    \item As a fully connected layer learns from the global representation of the network, we propose an idea of having fully connected layers as aggregated module which is to be proven effective in \cite{resnext}.
    \item We propose a method of having an multi-branch fully connected layers for which the squeezed layer passes important features before excitation to normal shape. This aggregated module not only has less complexity but enhance the performance than traditional aggregated modules as only important features are being fed as input and other features are omitted.
\end{itemize}

\section{Related Work}
The prior work related to the proposed model is seen in this section. From the introduction of ConvNets by YannLe Cunn, the convolutional neural network (CNN) have success in vision applications. Later multiple networks have been proposed once the researchers have access to better computational machines. This helped the researchers to contribute significant work in field network engineering. Architecture uses CNN for better performance have been seen from AlexNet \cite{alexnet}, ZFNet \cite{zfnet}, VGG \cite{vgg2014}, Resnet \cite{resnet}. The most relevant methods to the proposed architecture are multi branch convolutions and grouped convolutions.

During 2015, Inception created a wave in architectural design which achieves competitive performance with lesser computational complexity. It uses multi branch convolutions where the convolutions in branches are customized. It utilizes the image module to the maximum extent by adding multiple convolutional filters within a multi-branched structure. This won the ILSRVC in 2014 and reduced the parameters of previous best AlexNet from 60 million to 4 million. This multibranch convolutions later have been used in aggregated ResNet modules. Some of the notable architectures which emerged from the inception  having traits directly or indirectly are ResNeXt, Xception, Mobilenet \cite{mobilenet,xception}. The inception module shows the benefits of having deeper networks. Xception architectures have splitted the convolution operations within the inception module which makes the convolution operation much faster. From xception, Mobilenet also uses depthwise separable convolutions for all layers which have lesser computations and model size is comparable smaller in nature. 

Grouped convolutions distribute models over multiple GPUs. Alexnet uses the grouped convolutions to distribute the model over two GPUs. There's little evidence that this type of convolutions helped in increasing accuracy to the best of our knowledge.Channel wise convolution is a variant from grouped convolution with the total number of groups are equal to the number of channels.

Residual networks have established the new wave by proposing a solution of vanishing gradient. When going deeper architecture, model performance gets degraded as the learned representations are not being transformed to deep networks. This can be overcome by the residual module of resnet by giving shortcut connections which pass previous learned representations repeatedly at regular intervals. Batch normalization helps stabilize the learning by regulating inputs to the layers \cite{batchnorm}. This normalization works on the batch wise inputs. Residual networks have multiple variants as it can expand its depth even to 1000 layers and forms a testing base for researchers \cite{resnet1000}. 

The upgradation of residual networks have been going on for long time. The variants are discussed one by one. Biglittle resnet \cite{biglittlenet} which alters the residual network by creating two branches. One branch keeps the original residual structure intact called ’big’ and another branch called little focuses on convolution layers with smaller feature maps by allowing models to learn other patterns as well. Resnext \cite{resnext} proposed an aggregated residual module by creating a new dimension of cardinality. Resnext have multibranch convolution networks similar to the inception module but with the same topology. The multibranch is connected with concatenation of resulting output from all the residual branches and passed to the next layer. Resnext reduces theoretical complexity and also enhances the performance better than traditional stacked architecture as claimed by Saining \cite{resnext}. Wide residual network (WRN) \cite{wrns} is another form of residual network which has an expanded width of convolution layers. This makes the module learn more than traditional resnet and also passes the learned information via shortcut connections. This collectively enhances the performance of the module. Researchers proposed resnest which uses split attention networks \cite{resnest}. In this resnest, the module creates a multipath network with a combination of channel wise representations. This can be done by splitting the layer with 1x1 convolution and then followed by 3x3 convolution. Then everything is combined with a split attention block. SEnet \cite{senet} proposed a resent based squeeze and excitation module which increases accuracy by channel wise representations in the module. Highway networks have a gated mechanism for regulating shortcut connections \cite{highway}. 

There is a relation with different filters used within architectures as it increases the representation of the network by learning smaller patterns. Existing approaches include inception \cite{inception}, Pyramid network \cite{pyramid} have multiple topologies mixed to provide better performance.In xception \cite{xception} and mobilenets \cite{mobilenet}, the depth wise separable convolution implements depthwise and pointwise convolution. In depth-wise convolution, a single filter is applied on each input channel. Then the pointwise convolution is applied where the 1x1 kernel filter is applied on each value. The channel-wise convolution is part of separable convolution where in the channel wise the convolutions are completed before pointwise convolutions. 

These architectures do have reduced parameters when compared with similar performing architectures or non aggregated networks. This work is related to the squeeze operation which deliberately reduces the structure of the network. There are architectures like Structured transform networks \cite{structured}, Deep fried convnets \cite{deepfried}, Shufflenet \cite{shufflenet} which have considered computational aspects as well. Architectures have focused on reduced computation, small networks and some have effect on mobile based applications. There are other approaches as well to create smaller networks by shrinking compression methods like quantization of pretrained networks. The attention mechanism in CNN focuses on the most important parts of the image and neglect the irrelevant parts. This helps in understanding the complex scenes in an image effectively. They are typically used by combining softmax function or sigmoid function which acts like a gating mechanism. The attention mechanism is used in squeeze and excitation network within SE block which models channel-wise relationships. The similar approach has been taken in proposed SaE network with lightweight gating mechanism to focus on channel-wise relationship in the network.

Similar related architecture for our proposed approach is the ensemble method. Ensembles have multiple replicas of the same model which work in parallel for the same problem and results are chosen based on all the results combined. Ensemble models have better performance than single models as they predict better.

\section{Representation comparison of SEnet and SaEnet}
% Title
% make hay while the sun shines
% Make representations when 
Researchers claimed that having aggregated modules containing more than one convolution operation by branching the input is effective than having deeper network or wider layers \cite{resnet,wrns,resnext}. This increased cardinality operations irrespective of filter size being same or different have impact on accuracy. This results when the model has been learning better spatial representations including the convolutional layers in the aggregated modules. The proposed one have enough layers to learn when the important information is being given as input. The idea of learning global representation in between spatial learning is proposed by Squeeze and excitation networks. Squeeze operation enable the learned information from global receptive field accessible to following layers which enhance performance. Making the global information to be learned with increased cardinality will make the model to learn better than existing approaches.

\begin{figure}[hbt!]
    \centering
    \includegraphics[width=0.48\linewidth]{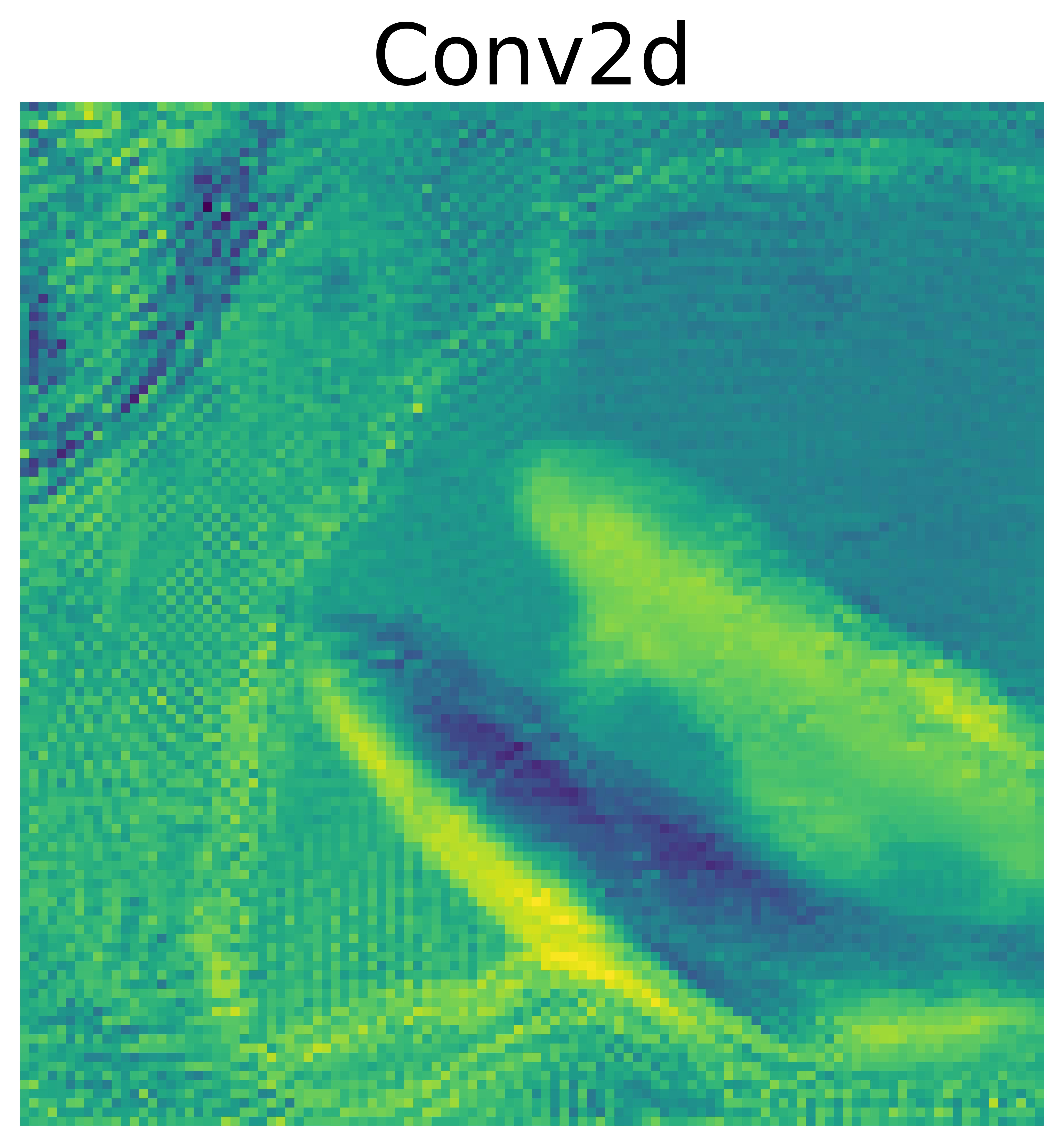}\hfill
    \includegraphics[width=.48\linewidth]{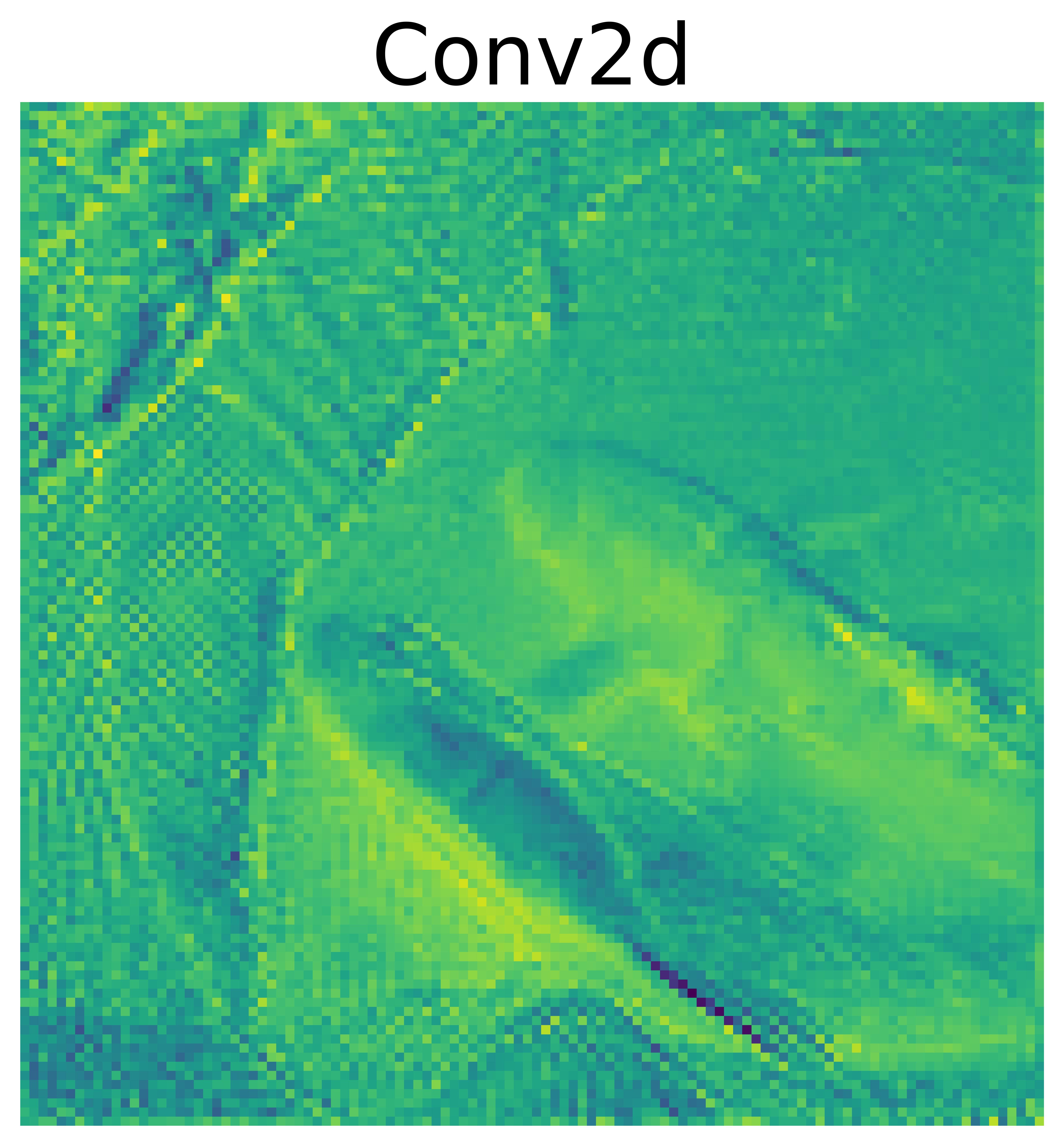}
    \caption{This explains the learned activation kernels for particular input dataset. We compare the learned values for the first convolution of SE resnet and the proposed SaE resnet models.}
    \label{fig:explainable}
\end{figure}

The images from figure \ref{fig:explainable} depict the learned representations for the first convolutional layer of both the SEnet and the proposed SaEnet. The learned representations have been taken after model is trained from scratch for 50 epochs with initial learning rate of 0.01 with the steps of 15 epochs the value is decayed with the rate of 0.1. This makes the model to learn in a better way. Due to the smallest computational power that we have, we are able to run it for 50 epochs. The images depicts that the proposed SaEnet learns more representations than SEnet as the consolidated information is passed on to the much more layers than SEnet.

\section{Methodology}
As discussed in the previous sections, the proposed network learns spatial representations from convolutional layers, channel-wise representations from squeeze and excitation layer and better channel-wise global representations from the aggregated layers. The proposed module is closely related to the ResNeXt which contains the aggregated molecule within the residual module. In resnext, researchers proposed the branched convolution contains the same layers with groups of size 32.

For explaining of the module better, we use the proposed approach with residual network \cite{resnet}. The residual module has become universal model for evaluation and bringing depth into consideration. Residual module contains shortcut connections which skip one or more layers. The layers are concatenated with the shortcut connection. The basic version of residual module, for an input x, the functions that are involved on altering input including batch normalization and dropout are indication with ‘F()’ the residual module is given by,

\begin{equation}
    Resnet = x+F(x)
\end{equation}

For aggregated module, the input alone is concatenated as they use branched convolutions and the resultant formula is given as,

\begin{equation}
    ResneXt = x + \sum_{}F(x)
\end{equation}

The core part evolved from the squeeze and excitation network. The Squeeze and excitation module consists of combination of two methods namely Squeeze and Excitation. Squeeze module contains the input to be squeezed with fully connected (FC) layers. The output of the convolutional layer is fed into the global average pooling layer to generate the channel wise input. Then the input is fed to FC layer with the reduction size. The excitation part of module consists of having FC layer without reduction to bring it back to original form. FC layer followed by scaling operation wherein the output is scaled by channelwise multiplication with the feature map. The final output is rescaled to its original shape. Squeeze and excitation operation for residual module is formulated as,

\begin{equation}
    SEnet = x + F(x \cdot Ex(Sq(x)))
\end{equation}

Wherein, ‘Sq’ function specifies the Squeeze operation involving FC with reduction size ‘r’. The 'Ex' operation is excitation operation which happens after ‘Sq’ to reshape the channelwise modified inputs to same shape without reduction. This is followed by a scaling operation with the input to bring its original form. This is concatenated with the input as it is in the residual module. 

Having the bigger groups in the squeezed format may increase the size and deviates the core idea of squeezed excitation. Thus we tested by having a cardinality of 4 which enhances itself as the core important features only are engaging with the excitation layer. This cardinality is enough to learn global representations better.

\begin{figure}
    \centering
    \includegraphics[width=0.9\linewidth,height=6cm]{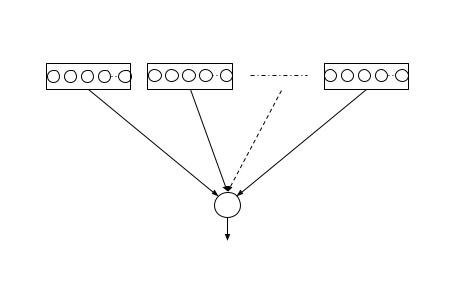}
    \caption{Figure represents the proposed aggregated fully connected layers within SaE net. This combines with concatenation at the end. The compressed conv layers after squeeze is fed as input to this module. The layers consists of reduction size 32 with the cardinality 4.}
    \label{fig:aggregated}
\end{figure}

In the squeeze operation, the FC layer with reduced size acts on the output of global average pooling. This conversion makes the important features to pass through the module and boosts the representational power of the network. We propose the increased cardinality of that FC layer. Having branched FC while reduction makes the model to learn more global representations in the network. The aggregated layers inside squeeze operation are concatenated and passed to the FC layer as shown in Figure \ref{fig:aggregated}. Then the output from FC is multiplied with input layer of the module for regaining the dimension. This final output is obtained by a scaling operation similar to SENet. This operation inside a residual module is depicted as follows,

\begin{equation}
    SaEnet (Proposed) = x + F(x \cdot Ex(\sum_{}Sq(x)))
\end{equation}

% \subsection{Squeeze and Excitation}
% The proposed module of squeeze and excitation is mathematically explained previously. The core operation in comparison with existing methods is discussed in this section. 

\section{Squeeze aggregated excitation resnet}

%\textbf{Resnet starts:}
As the proposed module is on the existing SENet, it have the same characteristics of SEnet. This module can be integrated in architectures. The standard architectures output layer can be fed directly to the proposed module. We implement the proposed module on residual network as it is the testing base for most of the models \cite{resnext,biglittlenet,senet} and for easy evaluation of experiments with existing architectures.

\begin{figure}
    \centering
    \includegraphics[width=0.95\linewidth,height=8cm]{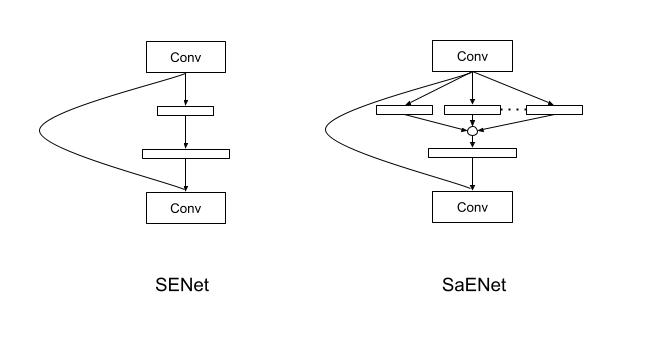}
    \caption{Figure represents the overall comparison between the squeeze and excitation resnet module with the proposed squeeze aggregated excitation resnet module. In figure, the small rectangle represents fully connected (FC) layers and the convolutional layers are represented as 'Conv'.The squeeze operation followed by excite or aggregated excite is always combined with the input so as to regain its original shape.}
    \label{fig:sevssae}
\end{figure}

The proposed SaE module is explained in detail on implementing in resnet architecture. The residual module contains a shorcut connection which appends after certain layers to pass the learned information of network. This helps in building deeper networks by avoiding vanishing gradient problem. The SaE module is incorporating module in residual module without changing the architecture. The purpose of choosing residual module is that it is passing information constantly each time and the learning of network can be made better if residual module learns better. Researchers implement the proposed modules on the residual module for this reason \cite{resnext,senet,biglittlenet}. The squeeze operation uses summation for all the branched FC layers followed by excitation operation. Figure \ref{fig:sevssae} shows the comparison between the SE module with the proposed module on resnet module. The indepth function of Squeeze and excitation is explained in the figure. 

% Squeeze : Global average pooling 
% Excitation: Two FC (reduction,dimensionality increasing) layers, Sigmoid+relu

The squeeze operation after the global average pooling layer for getting channel wise statistics. This is then fed to squeeze operation by shrinking the input. This is followed by excitation layer. The residual module consists of a repeating convolutional layer after certain layers forming a module. This type of modules repeat on a periodic basis for passing the learned gradients and not letting them vanish when going deeper. The residual network is structured with the base of VGG network \cite{vgg2014,resnet}.

\section{Implementation}
We implement the proposed method in comparison with state of the art architectures. We use two datasets for comparison CIFAR100 and Imagenet. On the datasets CIFAR 100 and Imagenet, we use input of size 224x224. We use H100 for testing the proposed method.  The proposed module can be incorporated with any existing module. This has similar characteristics to the SE module. We implement the proposed module on ResNet and compared with other architectures. The residual module have been implemented for deeper models, we implement that characteristic along with proposed SENeXt. The comparison between the residual network along with the proposed Squeeze aggregated excitation module on resnet and resnext is tabulated \ref{tab:se_design}

\begin{table*}[hbt]
    \centering
    \setlength{\tabcolsep}{15pt}
    \renewcommand{\arraystretch}{1.5}
    \begin{tabular}{c | c | c}
        \hline
        Resnet-50 & SaE resnet-50 & SaE resneXt-50 \\[3pt]
        \hline
        \multicolumn{3}{c}{conv, 7x7, 64, stride 2} \\
        \hline
        \multicolumn{3}{c}{max pool, 3x3, stride 2} \\
        \hline
        $\begin{bmatrix}
             conv,1x1,64 \\[3pt]
             conv,3x3,64 \\[3pt]
             conv,1x1,256 \\[3pt]
            \end{bmatrix} \times 3 $ & 
        
        $\begin{bmatrix}
             conv,1x1,64 \\[3pt]
             conv,3x3,64 \\[3pt]
             conv,1x1,256 \\[3pt]
             fc,[8,256] \times 4 \\[3pt]
        \end{bmatrix}  \times 3 $ &
        
        $\begin{bmatrix}
             conv,1x1,128 \\[3pt]
             conv,3x3,128 \ C=32 \\[3pt]
             conv,1x1,256 \\[3pt]
             fc,[8,256] \times 4 \\[3pt]
        \end{bmatrix}  \times 3 $\\
        \hline
        % second
        $\begin{bmatrix}
             conv,1x1,128 \\[3pt]
             conv,3x3,128 \\[3pt]
             conv,1x1,512 \\[3pt]
            \end{bmatrix} \times 4 $ & 
        
        $\begin{bmatrix}
             conv,1x1,128 \\[3pt]
             conv,3x3,128 \\[3pt]
             conv,1x1,512 \\[3pt]
             fc,[16,512] \times 4 \\[3pt]
        \end{bmatrix}  \times 4 $ &
        
        $\begin{bmatrix}
             conv,1x1,256 \\[3pt]
             conv,3x3,256 \ C=32 \\[3pt]
             conv,1x1,512 \\[3pt]
             fc,[16,512] \times 4 \\[3pt]
        \end{bmatrix}  \times 4 $\\
        \hline
        % third
        $\begin{bmatrix}
             conv,1x1,256 \\[3pt]
             conv,3x3,256 \\[3pt]
             conv,1x1,1024 \\[3pt]
            \end{bmatrix} \times 6 $ & 
        
        $\begin{bmatrix}
             conv,1x1,256 \\[3pt]
             conv,3x3,256 \\[3pt]
             conv,1x1,1024 \\[3pt]
             fc,[32,1024] \times 4 \\[3pt]
        \end{bmatrix}  \times 6 $ &
        
        $\begin{bmatrix}
             conv,1x1,512 \\[3pt]
             conv,3x3,512 \ C=32 \\[3pt]
             conv,1x1,1024 \\[3pt]
             fc,[32,1024] \times 4 \\[3pt]
        \end{bmatrix}  \times 6 $\\
        \hline
        %fourth
        $\begin{bmatrix}
             conv,1x1,512 \\[3pt]
             conv,3x3,512 \\[3pt]
             conv,1x1,2048 \\[3pt]
            \end{bmatrix} \times 3 $ & 
        
        $\begin{bmatrix}
             conv,1x1,512 \\[3pt]
             conv,3x3,512 \\[3pt]
             conv,1x1,2048 \\[3pt]
             fc,[64,2048] \times 4 \\[3pt]
        \end{bmatrix}  \times 3 $ &
        
        $\begin{bmatrix}
             conv,1x1,1024 \\[3pt]
             conv,3x3,1024 \ C=32 \\[3pt]
             conv,1x1,2048 \\[3pt]
             fc,[64,2048] \times 4 \\[3pt]
        \end{bmatrix}  \times 3 $\\
        \hline
        \multicolumn{3}{c}{global average pool, 1000-d fc, softmax} \\
        \hline
    \end{tabular}
    \caption{The shapes and operations along with groups (C) and aggregated FC layers depicted with the cardinality values. (\textbf{left}): Resnet-50, (\textbf{Middle}): SaE resnet-50 and (\textbf{Right}): SaE resneXt-50. The fc indicates the output dimension of the two fully connected layers in SaE network.}
    \label{tab:se_design}
\end{table*}

The above table represents the proposed Squeeze aggregated excitation module on residual network and aggregated residual network. The plain residual network is added on column right after the output size to compare with the proposed SaE resnet. Both the variants contain similar SaE characteristic along with its own model characteristics. All the convolutional layers are backed by the batch normalization technique. When compared to other normalization techniques, batch normalization provides a better by solving internal covariant shift. This normalization have better performance than other techniques. The normal SaE resnet starts with the convolutional layer which is followed by batch normalization. We also use the activation function relu in all the convolutional layers except the last layer which uses softmax as it is used for classification.

We explain the proposed model in comparison with vanilla residual network. Vanilla resnet is proposed for vanishing gradient problem. Later models proposed on top of resnet either having their own alterations \cite{biglittlenet} or making alterations in the module \cite{resnext,wrns,senet}. Alterations in the module is definitely being passed to the next layers using shortcuts. The proposed squeeze aggregated excitation (SaE) is an upgraded version of squeeze excitation (SE) module with extra cardinality parameter. SE module is proposed within the resnet module by squeezing and regaining to the same position using FC layers. The core part of SE module is altered with our proposed approach. Within the residual module, the squeeze module uses the global average pooling layer. We use the raw format by taking mean average of each tensor since the global average pooling layer is not in Pytorch.

The squeeze operation is followed by aggregation which increases cardinality. We also have a reduction size as SE module which allows the squeezed form of information can be forwarded to next layer. We use reduction of 32 for any model which reduces from the current input layer to SaE module. The squeezed output passes through the FC layer of reduced size. This layer is key layer after squeeze as this layer is receiving end of important information passed after squeeze. This have direct impact on the representational power of the network. We innovate by adding cardinality to this layer. We tried increasing the dense layers of the network. This increases the complexity of the network and also it doesnot have much effect when compared to multibranch module \cite{resnext}.

Thus we use multibranch FC layer for increasing representational power of the network. We use the branch value of 4 i.e., we use 4 fully connected layers which learns from the squeezed input. The value is chosen to have an impact on the learned representations at the same time it should be less complex to use even in resnext. We also tested the SaE module on aggregated resnet. This layer is combined using the concatenation as shown in figure \ref{fig:aggregated}. The concatenation passes all the layer outputs to the next layer. This FC layer brings the outputs to original shape as used by convolutional layers. The proposed SaE module is effective than SE module as it have a aggregated module which enhances the representations and the theoretical complexity is similar to existing SE module. 

\section{Experiments}
We experiment the proposed method on CIFAR-100, and modified Imagenet. Imagenet training images are modified and instructions are discussed in the upcoming section. For imagenet, we transform input data to batches of size 256 before training. We use Stochastic gradient descent (SGD) optimizer with the momentum of 0.9 and with the weight decay of 1e-4. This makes the network to learn slowly but effectively. We use the cross entropy loss function for all the datasets. Due to the computational constraint, we train all the models to be tested for 50 epochs on H100. Both the datasets follows same set of procedures and implemented in Pytorch. The initial learning rate for the model is 0.01 and it is decayed by the rate of 0.1 after each 15 steps. For resnext, the cardinality used is 32. 

\subsection{CIFAR-100}
We experiment on CIFAR 100 dataset. CIFAR-100 dataset contains 100 classifications with each class containing 6000 images. Among those images, we have 50000 for training and 10000 for testing purposes. We test on this dataset and the results are tabulated.

\begin{table}[hbt]
    \centering
    \begin{tabular}{c|c|c}
    \hline
    Models &  Top-1 & Top-5 \\
    \hline
    Resnet & 1.1000 & 4.9600 \\
    SE Resnet & 6.0200 & \textbf{10.2800} \\
    SaE Resnet & \textbf{6.0700} & 8.6500 \\ 
    \hline
    BLresnet &  \textbf{30.3600} & \textbf{59.8100} \\
    BLresnet + SE & 29.9300 & 59.6300 \\
    BLresnet + SaE & 29.5500 & 58.2400 \\
    \hline
    Aggregated resnet & 5.5500 & 9.0900 \\
    Aggregated resnet + SE & 5.7000 & 8.4600 \\
    Aggregated resnet + SaE & \textbf{5.7100} & \textbf{8.5300} \\
    \hline
    BL Aggregated resnet & 29.1200 & 58.4700 \\
    BL Aggregated resnet + SE module & \textbf{29.4600} & \textbf{58.8100}\\
    BL Aggregated resnet + SaE module & 28.3900 & 56.6600 \\
    \hline
    \end{tabular}
    \caption{Results obtained from experimenting on various models using CIFAR-100 dataset.}
    \label{tab:cifarres}
\end{table}

From the table \ref{tab:cifarres}, the squeeze and excitation and the proposed squeeze aggregated excitation is very close than vanilla resnet. In certain cases, like in SaE resnet, proposed network performed better on Top-1. And in Big little based aggregated resnet with SE module, both top-1 and top-5 is performed comparably better than other variants. We could see the basic version of SaE module outperforms the resnet and SE resnet.

\subsection{Imagenet results}
We have experimented on custom modified imagenet dataset. Due to the computational constrains we are unable to train the 1000 images per class. In Caltech 256 it consists a total of 30607 images wherein each class contains minimum of 80 images. The average images per class is 119. Thus, we consider a total of 250 images for imagenet which is 5 times more than the validation set of 50 per class. We used the same valideation set for evaluating the trained model. This modified dataset is used for testing imagenet with the above used data transformations.  
\begin{table}[]
    \centering
    \begin{tabular}{c|c|c}
        \hline
        Models & Top-1 & Top-5 \\
        \hline
        Resnet & 0.4000 & 0.6000 \\
        Resnet + SE & \textbf{0.5480} & 0.7740 \\
        Resnet + SaE & 0.4780 & \textbf{0.8280} \\
        \hline
        BL resnet & 21.8740 & 43.4680 \\
        BL resnet + SE & 22.1860 & 43.9180 \\
        BL resnet + SaE  & \textbf{22.2820} & \textbf{43.9680} \\
        \hline
        ResneXt & 0.1100 & 0.5380 \\
        ResneXt + SE & 0.3100 & 0.5860 \\
        ResneXt + SaE & \textbf{0.3292} & \textbf{0.7228} \\
        \hline
        BL resneXt & 22.8000 & 44.7600 \\
        BL resneXt + SE & 24.6640 & 47.7960 \\
        BL resneXt + SaE & \textbf{24.9200} & \textbf{47.9220} \\
        \hline
    \end{tabular}
    \caption{Results for the various models tested on modified Imagenet dataset. Train dataset consists of 250 images in the imagenet and validation data remains unaltered.}
    \label{tab:imagenetres}
\end{table}

The results for the experiments on models are tabulated \ref{tab:imagenetres}. From the obtained results, the proposed module have top-5 accuracy when compared to the vanilla resnet and SE resnet. Wherein, the SaE resnet achieves top-1 and top-5 on all the remaining experiment. When compared with biglittle resnet , resnext and in big little resnext, the proposed SaE module beats the plain network and squeeze excitation added networks.

\section{Conclusion}

In this paper, we propose the SaE module, a upgraded version of SE module, to improve the  representational power with the increased cardinality. The proposed model attached to residual network is subjected to extensive experiments which achieves better performance than the existing models. In addition to the architecture, we also introduce the aggregated fully connected layers for its global representation learning capabilities. We hope the combination of spatial, channel-wise and global representations in between network architecture proves useful for having better representations. Finally, the proposed network may be helpful in learning important features better in vision based related fields.

% \begin{center}
% \begin{tabular*}{c|c|c|c}
%  \hline
%  Output size & ResNet-50 & SxE-ResNet-50 & SxE-ResNeXt-50 \\
%  \hline
%  112 x 112 & \multicolumn{3}{c}{Conv} \\
%  \hline
%  cell7 & cell8 & cell9 \\
%  \hline
% \end{tabular*}
% \end{center}
{\small
\bibliographystyle{ieee_fullname}
\bibliography{PaperForReview}
}

\end{document}